\documentclass[runningheads]{llncs}
\usepackage{graphicx}
\usepackage{comment}
\usepackage{amsmath,amssymb} % define this before the line numbering.
\usepackage{color}

\usepackage{epsfig}
\usepackage{mathrsfs}
\usepackage{makecell}
\usepackage{float}
\usepackage{dsfont}
\usepackage{booktabs}

\newcommand{\etal}{et al.~}
\begin{document}
% \linenumbers
\pagestyle{headings}
\mainmatter
\def\ECCVSubNumber{3751}  % Insert your submission number here

\title{Piggyback GAN: Efficient Lifelong Learning for Image Conditioned Generation} % Replace with your title

\begin{comment}
\titlerunning{ECCV-20 submission ID \ECCVSubNumber} 
\authorrunning{ECCV-20 submission ID \ECCVSubNumber} 
\author{Anonymous ECCV submission}
\institute{Paper ID \ECCVSubNumber}
\end{comment}
%******************

%\begin{comment}
\titlerunning{Piggyback GAN}
\author{Mengyao Zhai \and Lei Chen \and Jiawei He \and Megha Nawhal \and \\
        Frederick Tung \and Greg Mori}
\authorrunning{M. Zhai et al.}
\institute{Simon Fraser University\\
\email{\{mzhai, chenleic, jha203, mnawhal, ftung\}@sfu.ca \; mori@cs.sfu.ca}}
%\end{comment}
%******************

%-------------------------------------------------------------------------
\makeatletter
\g@addto@macro\@maketitle{
  \begin{figure}[H]
  \setlength{\linewidth}{\textwidth}
  \setlength{\hsize}{\textwidth}
  \centering
  \includegraphics[width=\textwidth]{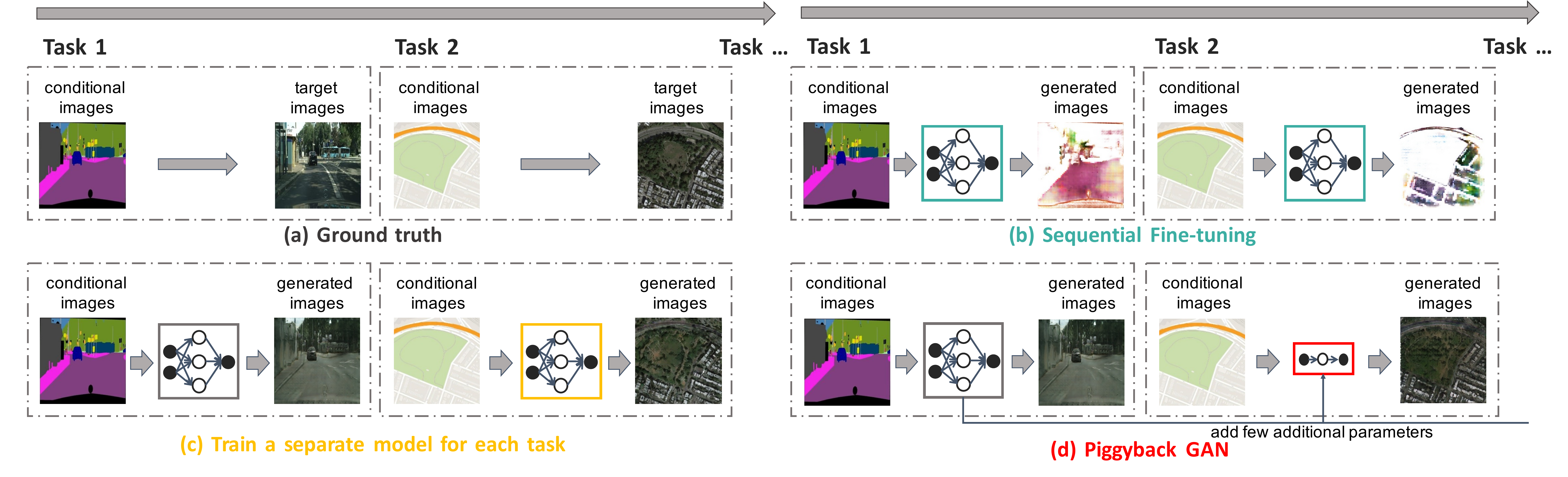}
\caption{\textbf{Lifelong learning of image-conditioned generation.} The goal of lifelong learning is to build a model capable of adapting to tasks that are encountered sequentially. Traditional fine-tuning methods are susceptible to catastrophic forgetting: when we add new tasks, the network forgets how to perform previous tasks (Figure~\ref{fig:intro} (b)). Storing a separate model for each task addresses catastrophic forgetting in an inefficient way as each set of parameters is only useful for a single task (Figure~\ref{fig:intro} (c)). Our Piggyback GAN achieves image-conditioned generation with high image quality on par with separate models at a lower parameter cost by efficiently utilizing stored parameters (Figure~\ref{fig:intro} (d)).}
  \label{fig:intro}
  \end{figure}
}
\makeatother
%-------------------------------------------------------------------------

\maketitle

\begin{abstract}
Humans accumulate knowledge in a lifelong fashion. Modern deep neural networks, on the other hand, are susceptible to catastrophic forgetting: when adapted to perform new tasks, they often fail to preserve their performance on previously learned tasks. Given a sequence of tasks, a naive approach addressing catastrophic forgetting is to train a separate standalone model for each task, which scales the total number of parameters drastically without efficiently utilizing previous models. In contrast, we propose a parameter efficient framework, Piggyback GAN, which learns the current task by building a set of convolutional and deconvolutional filters that are factorized into filters of the models trained on previous tasks. For the current task, our model achieves high generation quality on par with a standalone model at a lower number of parameters. For previous tasks, our model can also preserve generation quality since the filters for previous tasks are not altered. We validate Piggyback GAN on various image-conditioned generation tasks across different domains, and provide qualitative and quantitative results to show that the proposed approach can address catastrophic forgetting effectively and efficiently.
% piggyback filters from linear combinations of 

\end{abstract}

\section{Introduction}
% Writing should highlight:\\
% \noindent
% -- generic: in terms of various conditions\\
% -- efficient: weight-reuse\\
% -- maintain image quality: for the new task\\
% -- no quality deduction: for the old task\\

%It is important to foster a love for lifelong learning. We human are all 
Humans are lifelong learners: in order to function effectively day-to-day, we acquire and accumulate knowledge throughout our lives. The accumulated knowledge makes us efficient and versatile when we encounter new tasks. In contrast to human learning, modern neural network based learning algorithms usually fail to remember knowledge acquired from previous tasks when adapting to a new task (see Figure~\ref{fig:intro} (b)). In other words, it is difficult to generalize once a model is trained on a task. This is the well-known phenomenon of \textit{catastrophic forgetting}~\cite{mccloskey1989catastrophic}. Recent efforts~\cite{shmelkov2017incremental,li2017learning,chaudhry2018efficient,aljundi2018memory} have demonstrated how discriminative models can continually learn a sequence of tasks. Despite the success of these efforts for discriminative models, lifelong learning for generative models remains a challenging and under-explored area.

Lifelong learning methods for discriminative models cannot be directly applied to generative models due to their intrinsic differences. \textit{First}, it is well known that in classification tasks,  the intermediate convolutional layers in deep neural networks are capable of providing generic features.
These features can easily be reused by other models with varying and different classification goals. For generative models, the possibility of such reuse for new generative tasks has not been previously explored, to the best of our knowledge. \textit{Second}, different from discriminative models, the output space of generative models is usually continuous, making it more challenging for the model to maintain the generation quality along the training of a sequence of tasks. \textit{Third}, there could be conflicts between tasks under the generative setting. For discriminative models, it rarely happens that one image has different labels (appears in different tasks). However, for generative models, it is quite common that we want to translate the same image to different domains for different tasks~\cite{isola2016pix2pix,gatys2016image,zhu2017unpaired,choi2018stargan}.

One of the naive approaches to address catastrophic forgetting for generative models is to train a model for each task separately (see Figure~\ref{fig:intro} (c)). Unfortunately, this approach is not scalable in general: as new tasks are added, the storage requirements grow drastically. More importantly, setting the trained model aside without exploiting the benefit it can potentially provide facing a new task would be an inefficient use of resources. 
Recent works~\cite{zhai2019lifelong,wu2018memory} have shown promising results on lifelong learning for generative models, but it is also revealed that image quality degradation and artifacts transfer from old to new tasks are inevitable. Therefore, it is valuable to have a continual learning framework designed that (1) is more parameter efficient, (2) preserves the generation quality of both current and previously learned tasks, and (3) can enable various conditional generation tasks across different domains.

In this paper, we introduce a generic continual learning framework \textit{Piggyback GAN} that can perform various conditional image generation tasks across different domains (see Figure~\ref{fig:intro} (d)). The proposed approach addresses the catastrophic forgetting problem in generative continual learning models. Specifically, \textit{Piggyback GAN} maintains a \textit{filter bank}, the filters in which come from convolution and deconvolution layers of models trained on previous tasks. Facing a new task, Piggyback GAN learns to perform this task by reusing the filters in the filter bank by building a set of \textit{piggyback filters} which are factorized into filters from the filter bank. Piggyback GAN also maintains a small portion of unconstrained filters for each task to ensure high-quality generation. Once the new task is learned, the unconstrained filters are appended to the filter bank to facilitate the training of subsequent tasks. Since the filters for the old task have not been altered, Piggyback GAN is capable of keeping the exact knowledge of previous tasks without forgetting any details. 

To summarize, our contributions are as follows. We propose a continual learning framework for conditional image generation that is (1) \textit{efficient}, learning to perform a new task by ``piggybacking" on models trained on previous tasks and reusing the filters from previous models, (2) \textit{quality-preserving}, maintaining the generation quality of current task and ensuring no quality degradation for previous tasks while saving more parameters, and (3) \textit{generic}, enabling various conditional generation tasks across different domains. To the best of our knowledge, we are the first to make these contributions for generative continual learning models. We validate the effectiveness of our approach under two settings: (1) paired image-conditioned generation, and (2) unpaired image-conditioned generation.
Extensive qualitative and quantitative comparisons with state-of-the-art models are carried out
to illustrate the capability and efficiency of our proposed framework to learn new generation tasks without the catastrophic forgetting of previous tasks. 
% To the best of our knowledge, we are the first to make these contributions for continual learning generative frameworks.

% \textcolor{red}{Eric:Paragraph 3:
% The beginning of the current paragraph seems a bit odd, since it's a general statement for lifelong learning models. Following the flow from paragraph 2, it might be better if we directly start with previous works on lifelong learning for generative models (might even be a good idea to merge to paragraph 2.) Then state the weakness of previous works.
% }

% \textcolor{red}{Eric:Paragraph 4:
% An IMPORTANT paragraph introducing our model. Need to clarify what we can do that previous works cannot. The word ``filters" appears too early here. Might need a more general word here since the concept of filter is not properly introduced yet.
% }

% \textcolor{red}{Eric:Paragraph 5:
% Seems fine, just needs to refine wording. Also, might be a good idea to mention the three problems in paragraph 2 again. Just to say we overcame or alleviated these problems.
%}

% \textcolor{leipink}{Lei: In the analogy to human beings we might also emphasis something like human beings learn new stuff by linking them to old knowledge, motivating our approach of filter reusing. Some comparisons with discriminative models seem a bit less rigorous.}
\section{Related Work}
% \textcolor{mzhai}{Mengyao: Thanks @Fred for the nice section that covers a wide range of papers, but we might need to shrink the number of citations in the discriminative side. For generative part, I don't think we have many papers to cite. The huge difference in the number of cited papers in these two parts might cause reviewers focusing on the wrong direction. \\}
% \textcolor{ft}{Fred: Removed 8 references from the first paragraph of 'parameter-efficient network learning' - I hope that is the paragraph you mean. \\}
% \textcolor{mzhai}{Mengyao: Exactly, thanks very much! \\}

Our work intersects two previously disjoint lines of research: lifelong learning of generative models, and parameter-efficient network learning.

\textbf{Lifelong learning of generative models.}
For discriminative tasks, e.g.\ classification, many works have been proposed recently for solving the problem of catastrophic forgetting. Knowledge distillation based approaches \cite{shmelkov2017incremental,li2017learning,castro2018end} work by minimizing the discrepancy between the output of the old and new network. Regularization-based approaches \cite{kirkpatrick2017overcoming,chaudhry2018riemannian,aljundi2018memory,chaudhry2018efficient,lopez2017gradient} regularize the network parameters when learning new tasks. Task-based approaches \cite{Serr2018OvercomingCF,aljundi2017expert,rusuetal2016} adopt task-specific modules to learn each task.

For generative tasks, lifelong learning is an underexplored area and relatively less work studies the problem of catastrophic forgetting. Continual generative modeling was first introduced by Seff \etal \cite{seff2017continual}, which incorporated the idea of \textit{Elastic Weight Consolidation (EWC)}~\cite{kirkpatrick2017overcoming} into the lifelong learning for GANs. The idea of \textit{memory replay}, in which images generated from a model trained on previous tasks are combined with the training images for the current task to allow for training of multiple tasks, is well explored by Wu \etal \cite{wu2018memory} for label-conditioned image generation. However, this approach is limited to label-conditioned image generation and is not applicable
for image-conditioned image generation since no ground-truth conditional inputs of previous tasks are provided. \textit{EWC}~\cite{kirkpatrick2017overcoming} has been adapted from classification tasks to generative tasks of label-conditioned image generation~\cite{seff2017continual,wu2018memory}, but they present limited capability in both remembering previous tasks and generating realistic images. \textit{Lifelong GAN}~\cite{zhai2019lifelong} is a generic knowledge-distillation based approach for conditioned image generation in lifelong learning. However, the image quality of generated images of previous tasks keeps decreasing when new tasks are learned. Moreover, all approaches mentioned above fail in the scenario when there are conflicts in the input-output space across tasks. To the best of our knowledge, Piggyback GAN is the first method that can preserve the generation quality of previous tasks, and enables various conditional generation tasks across different domains.

\textbf{Parameter-efficient network learning.} 
In recent years, there has been increasing interest in better aligning the resource consumption of neural networks with the computational constraints of the platforms on which they will be deployed \cite{chenetal2018,wuetal2019,yangetal2018}. One important avenue of active research is parameter efficiency. Trained neural networks are typically highly over-parameterized \cite{deniletal2013}. There are many effective strategies for improving the parameter efficiency of trained networks: for example, pruning algorithms learn to remove redundant or less important parameters \cite{heetal2019,yuetal2018}, and quantization algorithms learn to represent parameters using fewer bits while maintaining task performance \cite{jacobetal2018,jungetal2019,zhangetal2018}. Parameter efficiency can also be encouraged during training via architecture design \cite{maetal2018,sandleretal2018}, sparsity inducing priors \cite{tartaglioneetal2018}, resource re-allocation \cite{qiaoetal2019}, or knowledge distillation from a teacher network \cite{tungmori2019,zagoruykokomodakis2017}. 
Our work is orthogonal to these methods, which may be applied on top of Piggyback GAN.

The focus of this work is parameter-efficient lifelong learning: we aim to leverage weight reuse and adaptation to improve parameter efficiency when extending a trained network to new tasks, while avoiding the catastrophic forgetting of previous tasks. Several approaches have been explored for parameter-efficient lifelong learning in a discriminative setting. For example, progressive networks \cite{rusuetal2016} train a new network for each task and transfer knowledge from previous tasks by learning lateral connections, a linear combination of layer activations from previous tasks is computed and added to the inputs to the corresponding layer of the new task. While our approach computes a linear combination of filters to construct new filters and uses them the same as normal filters. Rebuffi \etal \cite{rebuffietal2018} aims to train a universal vector that are shared among all domains providing generic filters for all domains, and task-specific adapter modules are added to adjust the trained network to new tasks. While we are working on solving a sequence of tasks, the filters trained for the initial task are hardly generic enough for later tasks. Inspired by binary neural networks, Mallya \etal \cite{mallyaetal2018} introduced the piggyback masking method (which inspired the naming of Piggyback GAN). This method learns a binary mask over the network parameters for each new task. Tasks share the same backbone parameters but differ in the parameters that are enabled. Multi-task attention networks 
\cite{liuetal2019} learn soft attention modules for each new task. Task-specific features are obtained by scaling features with an input-dependent soft attention mask.
To the best of our knowledge, Piggyback GAN is the first method for parameter-efficient lifelong learning in a generative setting.

\section{Method}

\begin{figure*}[ht]
\begin{center}
  \includegraphics[width=\textwidth]{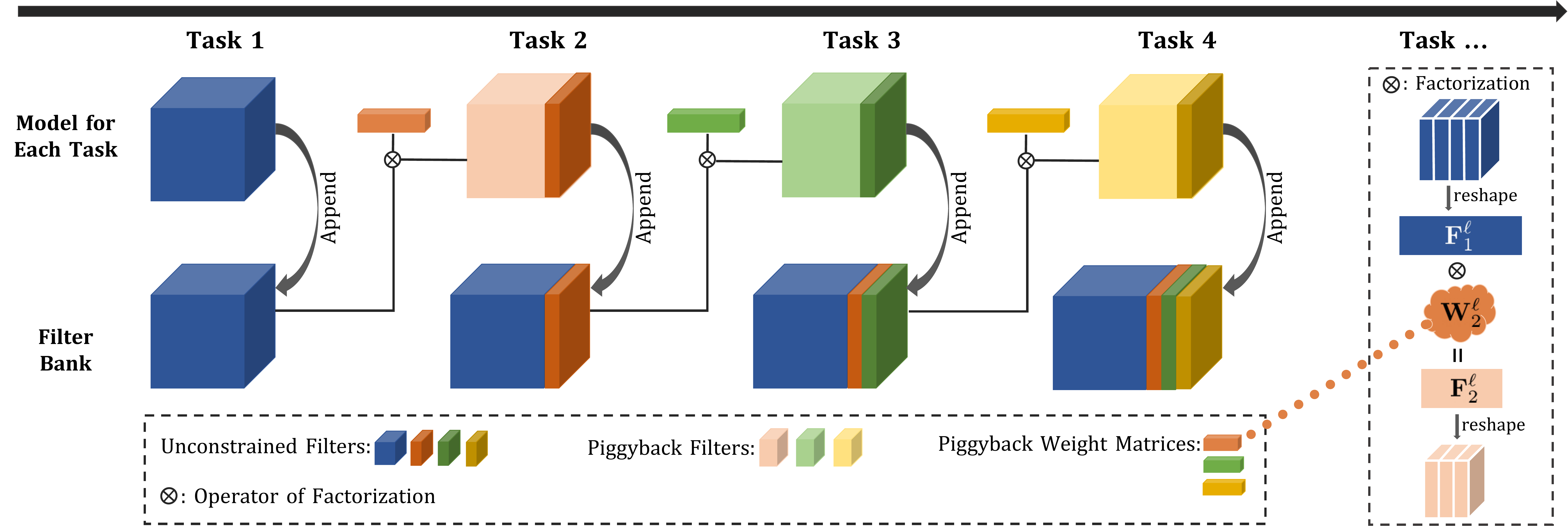}
\end{center}
\caption{\textbf{Overview of Piggyback GAN}. Piggyback GAN maintains a \textit{filter bank} for composing convolutional filters to be used for the next task. Given a new task, most filters in the model are trained as \textit{piggyback filters} that are factorized into filters in the filter bank. The remaining small portion of filters are learned without constraints and appended to the filter bank to benefit the training of subsequent tasks.}
\label{fig:overview}
\end{figure*}

In this paper, we study the problem of lifelong learning for generative models. The overall goal is to learn the model for a given sequence of generation tasks $\{T_i\}_{i=1}^N$ under the assumption that the learning process has access to only one task at a time and training data for each task is accessible only once. 

Assume that we have a learnt model $M$ trained to perform the first task $T_1$. To proceed with the new task $T_2$, one naive approach is to adapt the learnt model $M$ to the new task by fine-tuning its parameters. This approach suffers from the problem of catastrophic forgetting of task $T_1$. Another naive approach is to learn a separate standalone model $M_i$ for each new task $T_i$ and store all learnt models $\{M_i\}_{i=1}^N$, which drastically scales up total number of parameters without utilizing previous models. 

The key idea of this paper is to build a parameter efficient lifelong generative model by ``piggybacking" on existing models and reusing the filters from models trained on previous tasks while maintaining the generation quality to be similar to a single standalone model for all tasks. We achieve this by maintaining a filter bank, and compose the filters for new tasks mostly by factorizing it into filters in the filter bank, with a small portion of filters learnt without constraints. Once the new task is learnt, the filter bank is expanded by adding those unconstrained filters.

\subsection{Piggyback Filter Learning}
\label{sec:piggyback-filter}
For lifelong learning of generative models, given an initial task $T_1$ and an upcoming new task $T_2$, there could be cases where $T_1$ and $T_2$ share similar inputs but different outputs, e.g. $T_1$ is \textit{Photo $\rightarrow$ Monet Paintings} and $T_2$ is \textit{Photo $\rightarrow$ Ukiyo-e Paintings}. These conflicts in input-output space do not occur when each task is trained separately. However, as already mentioned, learning a separate model for each task and storing all learnt models is memory consuming and does not utilize previous models. 

For this scenario, the knowledge in the filters from the model trained for $T_1$ could provide valuable information for the training of $T_2$. 
For images from domains that are less visually similar, it is well known that in discriminative tasks, the generaliziblity of convolutional neural networks helps maintain its capability in previously unseen domains~\cite{mallyaetal2018,rebuffietal2018}. Such evidence suggests that despite the difference in fine details, certain general patterns that convey semantics could still be covered in previous filter space.
Inspired by these observations, we propose to utilize previously trained filters to facilitate the training of new tasks. In our experiments, we do find that previous filters make steady contribution to learning new tasks regardless of the domain difference across tasks.

Consider a generator network $G_1$ learnt for initial task $T_1$ consisting of $L$ layers of filters $\{F_1^{\ell}\}_{\ell=1}^L$ where $\ell$ denotes the index of layers. For the ${\ell}^{th}$ layer, let the kernel size be $s_w^{\ell} \times s_h^{\ell}$, number of input channels be $c_{in}^{\ell}$ and number of output channels be $c_{out}^{\ell}$. Then $F^{\ell}_1$ 
is reshaped from 4D into 2D of size $s_w^{\ell} \times s_h^{\ell} \times c_{in}^{\ell}$ by $c_{out}^{\ell}$, which we denote by $\mathcal{R}(F^{\ell}_1)$. For the upcoming task $T_2$, we learn the filters $F^{\ell}_2$ through factorization operation over filters of the corresponding layers in task $T_1$, namely

\begin{equation}
\begin{split}
    F^{\ell}_{2} \, = \, \mathcal{R}^{-1}(\mathcal{R}(F^{\ell}_{1}) \otimes W^{\ell}_2),
\end{split}
\end{equation}
where $\otimes$ denotes the standard matrix multiplication operation and $\mathcal{R}^{-1}$ denotes the inverse of reshape operation from 2D back to 4D. The derived filters $F^{\ell}_{2}$ is denoted as \textit{piggyback filters} and the corresponding learnable weights $W^{\ell}_2 \in \mathbb{R}^{c_{out}^{\ell} \times c_{out}^{\ell}}$ is denoted as \textit{piggyback weight matrix}. The resulting $F^{\ell}_{2} \in \mathbb{R}^{s_w^{\ell} \times s_h^{\ell} \times c_{in}^{\ell} \times c_{out}^{\ell}}$ is of the same size as $F^{\ell}_{1}$. A bias term could be added to the factorized filters to adjust the final output.

Therefore, by constructing factorized filters, the number of parameters required to be learnt for this layer is reduced from $s_w^{\ell} \times s_h^{\ell} \times c_{in}^{\ell} \times c_{out}^{\ell}$ to only $c_{out}^{\ell} \times c_{out}^{\ell}$. 

\subsection{Unconstrained Filter Learning}
\label{sec:unconstrained-filter}
Parameterizing a new task completely by making use of filters in the initial tasks, though saving substantial storage, cannot capture the key differences between tasks. The filters derived from previous tasks may only characterize certain attributes and may not generalize well to the new task, resulting in poor quality of generation details for the new model. Learning a small number of unconstrained filters gives the model greater flexibility in adapting to the new task. Moreover, learning these unconstrained filters could learn to generate patterns that do not exist in previous tasks, thereby, increasing the power of the model and helping the training of later tasks.

Specifically, for each layer in the generator, we allocate a small portion of unconstrained filters which are learned freely without the constraint to be constructed from previous filters. With a total of $c^{\ell}_{out}$ filters in $F^{\ell}_2$, consider the proportion of unconstrained filters to be $\lambda \, (0 < \lambda \leq 1)$\footnote{$\lambda c^{\ell}_{out}$ could be rounded to the nearest integer. Or $\lambda$ could be chosen to make $\lambda c^{\ell}_{out}$ an integer.}, then we have $\lambda c^{\ell}_{out}$ unconstrained filters and $(1-\lambda) c^{\ell}_{out}$ piggyback filters, and the size of \textit{piggyback weight matrix} $W^{\ell}_2$ has changed to $c_{out}^{\ell} \times (1-\lambda)c_{out}^{\ell}$. Let the unconstrained filters at the $l^{th}$ layer of the $n^{th}$ task $T_n$ be $F_n^{u,\ell}$, and piggyback filters at the $l^{th}$ layer of the $n^{th}$ task $T_n$ be $F_n^{p,\ell}$. 

For task $T_1$, all filters are unconstrained filters, namely
\begin{equation}
    F^{\ell}_{1} \, = \, F^{u,\ell}_{1}.
\end{equation}

For task $T_2$, we re-define the filters in $F^2_l$ to be the concatenations of unconstrained filters $F_2^{u,\ell}$ and piggyback filters $F_2^{p,\ell}$, namely $F_2^{\ell}$ is formulated as

\begin{equation}
\begin{split}
    F^{\ell}_{2} \, & = \, [F_2^{u,\ell}, F_2^{p,\ell}] \\
              & = \, [F_2^{u,\ell}, \mathcal{R}^{-1}(\mathcal{R}(F^{u,\ell}_{1}) \otimes W^{\ell}_2)],
\end{split}
\end{equation}
where $F_2^{u,\ell} \in \mathbb{R}^{s_w^{\ell} \times s_h^{\ell} \times c_{in}^{\ell} \times \lambda c_{out}^{\ell}}$, $F_2^{p,\ell} \in \mathbb{R}^{s_w^{\ell} \times s_h^{\ell} \times c_{in}^{\ell} \times (1-\lambda) c_{out}^{\ell}}$ and the resulting $F^{\ell}_{2} \in \mathbb{R}^{s_w^{\ell} \times s_h^{\ell} \times c_{in}^{\ell} \times c_{out}^{\ell}}$ is of the same size as $F^{\ell}_{1}$.

\subsection{Expanding Filter Bank}
\label{sec:filter-bank}

The introduction of unconstrained filters learnt from task $T_2$ brings extra storage requirement and it would be a waste if we set it aside when learning the filters for the upcoming new tasks. Therefore, instead of setting it aside, we construct the piggyback filters for task $T_3$ by making use of the unconstrained filters from both $T_1$ and  $T_2$.

We refer to the full set of unconstrained filters as the \textit{filter bank} which will be expanded every time new unconstrained filters are learnt for a new task. Expanding the \textit{filter bank} could encode more diverse patterns that were not captured or did not exist in previous tasks. By doing so, the model increases the modality of the learned filters and provides more useful information for the subsequent tasks. 

When learning the task $T_3$, the filter bank comprises filters corresponding to the task $T_1$ and $T_2$ and is given by $F^{u,\ell}_{1}$ and $F^{u,\ell}_{2}$, resulting in the size of the filter bank $s_w^{\ell} \times s_h^{\ell} \times c_{in}^{\ell} \times (1+\lambda) c_{out}^{\ell}$. Similarly, for task $T_3$ consider the proportion of unconstrained filters to be same as $T_2$, i.e. $\lambda$, the size of \textit{piggyback weight matrix} $W^{\ell}_3$ would be $(1+\lambda)c_{out}^{\ell} \times (1-\lambda)c_{out}^{\ell}$. We construct $F^{\ell}_{3}$ as the concatenation of unconstrained filters $F^{u,\ell}_{3}$ and piggyback filters $F^{p,\ell}_{3}$ that are constructed from all filters in the \textit{filter bank}, namely $F_3^{\ell}$ is formulated as

\begin{equation}
\begin{split}
    F^{\ell}_{3} \, & = \, [F^{u,\ell}_{3}, F^{p,\ell}_{3}] \\
               & = \, [F^{u,\ell}_{3}, \mathcal{R}^{-1}(\mathcal{R}([F^{u,\ell}_{1}, F^{u,\ell}_{2}]) \otimes W^{\ell}_3) \,],
\end{split}
\end{equation}
where $F_3^{u,\ell} \in \mathbb{R}^{s_w^{\ell} \times s_h^{\ell} \times c_{in}^{\ell} \times \lambda c_{out}^{\ell}}$,  $F_3^{p,\ell} \in \mathbb{R}^{s_w^{\ell} \times s_h^{\ell} \times c_{in}^{\ell} \times (1-\lambda) c_{out}^{\ell}}$ and the resulting $F^{\ell}_{3} \in \mathbb{R}^{s_w^{\ell} \times s_h^{\ell} \times c_{in}^{\ell} \times c_{out}^{\ell}}$ is of the same size as $F^{\ell}_{1}$ and $F^{\ell}_{2}$.

After task $T_3$ is learnt, the \textit{filter bank} is expanded as
$[F^{u,\ell}_{1}, F^{u,\ell}_{2}, F^{u,\ell}_{3}]$, whose size is $s_w^{\ell} \times s_h^{\ell} \times c_{in}^{\ell} \times (1+2\lambda) c_{out}^{\ell}$. 

To summarize, when learning task $T_{n}$, $F^{\ell}_{n}$ could be written as
\begin{equation}
\begin{split}
    F^{\ell}_{n} \, & = \, [F^{u,\ell}_{n}, F^{p,\ell}_{n}] \\
               & = \, [F^{u,\ell}_{n}, \mathcal{R}^{-1}(\mathcal{R}([F^{u,\ell}_{1}, F^{u,\ell}_{2},...,F^{u,\ell}_{n-1}]) \otimes W^{\ell}_n) \,].
\end{split}
\end{equation}

After task $T_{n}$ is learnt, the \textit{filter bank} is expanded as $[F^{u,\ell}_{1}, F^{u,\ell}_{2}, ..., F^{u,\ell}_{n}]$, whose size is $s_w^{\ell} \times s_h^{\ell} \times c_{in}^{\ell} \times (1+(n-1)\lambda) c_{out}^{\ell}$, and the size of \textit{piggyback weight matrix} $W^{\ell}_{n+1}$ for task $T_{n+1}$ is $(1+(n-1)\lambda)c_{out}^{\ell} \times (1-\lambda)c_{out}^{\ell}$. It should be noted that the weights of filters in the \textit{filter bank} remain fixed along the whole learning process. 

\subsection{Learning Piggyback GAN}
We explore two conditional generation scenarios in this paper: 
(1) \textit{paired image-conditioned generation}, in which training data contains $M$ pairs of samples \textit{$\{(a_i, b_i)\}_{i=1}^{M}$}, where \textit{$\{a_i\}_{i=1}^{M}$} represent conditional images, \textit{$\{b_i\}_{i=1}^{M}$} represent target images, and correspondence between $a_i$ and $b_i$ exists, namely for any conditional image $a_i$ the corresponding target image $b_i$ is also provided; (2) \textit{unpaired image-conditioned generation}, in which training data contains images from two domains $A$ and $B$, namely images \textit{$\{a_i\}_{i=1}^{M_a} \in A$} and images \textit{$\{b_i\}_{i=1}^{M_b} \in B$}, and correspondence between $a_i$ and $b_i$ does not exist, namely for any conditional image $a_i$ the corresponding target image $b_i$ is not provided.
% information like which $a_i$ matches which $b_i$ is not provided.

For both conditional generation scenarios, given a sequence of tasks and a state-of-the-art GAN model, we construct the convolutional and deconvolutional filters in the generator as described in Sec.~\ref{sec:piggyback-filter}, Sec.~\ref{sec:unconstrained-filter} and Sec.~\ref{sec:filter-bank}. The derived filters is updated by the standard gradient descent algorithm and the overall Piggyback GAN model is trained in the same way as any other existing generative models by adopting the desired learning objective for each task.

% \textcolor{mzhai}{Mengyao: Probably don't need the following equations. Could delete later if unnecessary.}
% \begin{equation}
% \begin{split}
%     \mathcal{L}_{\mathrm{GAN}}(G_n,D_n^Y) = \mathbb{E}_{y}[log D_n^Y (y)] + \mathbb{E}_{x}(x)[log (1-D_n^Y (G_n(x))] \\
% \end{split}
% \end{equation}

% \begin{equation}
% \begin{split}
%     \mathcal{L}_{\mathrm{p-GAN}}(G_n,D_n^Y) = \mathcal{L}_{\mathrm{GAN}}(G_n,D_n^Y) + \mathbb{E}_{x }||y-G_n(x)||_1
% \end{split}
% \end{equation}

% For learning the $n^{th}$ unpaired image-to-image translation task $X \leftrightarrow Y$, let the generator be $G_n$, the discriminator for domain $X$ be $D_n^X$ and discriminator for domain $Y$ be $D_n^Y$, the learning objective is:

% \begin{equation}
% \begin{split}
%     &\mathcal{L}_{\mathrm{u-GAN}}(G_n^{X \rightarrow Y},G_n^{Y \rightarrow X},D_n^X,D_n^Y) = \\ &\mathcal{L}_{\mathrm{GAN}}(G_n^{X \rightarrow Y},D_n^Y) + \mathcal{L}_{\mathrm{GAN}}(G_n^{Y \rightarrow X},D_n^X) + \\
%     & \mathbb{E}_{x}[||x - G_n^{X \rightarrow Y}(G_n^{Y \rightarrow X}(x))||_1]+ \mathbb{E}_{y}[||y - G_n^{Y \rightarrow X}(G_n^{X \rightarrow Y}(y))||_1]
% \end{split}
% \end{equation}
\section{Experiments}
We evaluate Piggyback GAN under two settings: (1) paired image-conditioned generation, and (2) unpaired image-conditioned generation. We first conduct an ablation study on the piggyback filters and unconstrained filters. We also demonstrate the generalization ability of our model by having same $T_2$
following different $T_1$. Finally, we compare our model with state-of-the-art approach~\cite{mallyaetal2018} proposed for discriminative models (i.e., classification models), which also shares the idea of ``piggybacking" on previously trained models by reusing the filters. Different from our approach, \cite{mallyaetal2018} learns a binary mask applying on the filters of a base network for each new task. 

\noindent \textbf{Training Details.} All the sequential generation models are trained on images of size $256 \times 256$. We use the Tensorflow~\cite{abadi2016tensorflow} framework with Adam optimizer~\cite{kingma2014adam}. We set the parameters $\lambda=\frac{1}{4}$ for all experiments. For paired image-conditioned generation, we use UNet architecture~\cite{isola2016pix2pix} and the last layer is set to be task-specific (a task-specific layer contains only unconstrained filters) for our approach and all baselines. For unpaired image-conditioned generation, we adopt the architecture~\cite{johnson2016perceptual,zhu2017unpaired} which have shown impressive results for neural style transfer, and last two layers are set to be task-specific for our approach and all baselines. For both conditional generative scenarios, bias terms are used to adjust the output after factorization.

\noindent \textbf{Baseline Models.} We compare Piggyback GAN to the following baseline models: (a) \textit{Full}: The model is trained on single task, which could be treated as the ``upper bound" for all approaches. (b) \textit{$\frac{1}{4}$Full} and \textit{$\frac{1}{2}$Full}: The model is trained on single task, and $\frac{1}{4}$ or $\frac{1}{2}$ number of filters used in baseline \textit{Full} is used. (c) \textit{Pure Factorization (PF)}: Model is trained with piggyback filters that are purely constructed from previously trained filters. (d) \textit{Sequential Fine-tuning (SFT)}: The model is fine-tuned in a sequential manner, with parameters initialized from the model trained/fine-tuned on the previous task. (e)\textit{~\cite{mallyaetal2018}}: a state-of-the-art approach for discriminative model (i.e. classification model) which reuses the filters by learning and applying a binary mask on the filters of a base network for each new task.

\noindent \textbf{Quantitative Metrics.} In this work, we use two metrics \textit{Acc} and \textit{Fréchet Inception Distance (FID)~\cite{heusel2017gans}} to validate the quality of the generated data. \textit{Acc} is the accuracy of the classifier network trained on real images and evaluated on generated images (higher Acc indicates better generation results). \textit{FID} is an extensively used metric to compare the statistics of generated samples to samples from a real dataset. We use this metric to quantitatively evaluate the quality of generated images (lower FID indicates higher generation quality). 
% We use this metric to measure how well each approach learns the new task without forget the old ones

\subsection{Paired Image-conditioned Generation}
We first demonstrate the effectiveness of Piggyback GAN on 4 tasks of paired image-conditioned generation, which are all image-to-image translations, on challenging domains and datasets with large variation across different modalities~\cite{cordts2016cityscapes,isola2016pix2pix,tylevcek2013spatial,yu2014fine}.
The first task is \textit{semantic labels} $\rightarrow$ \textit{street photos}, the second task is  \textit{maps} $\rightarrow$ \textit{aerial photos}, the third task is \textit{segmentations} $\rightarrow$ \textit{facades}, and the fourth task is \textit{edges} $\rightarrow$ \textit{handbag photos}.

\begin{figure}[h!]
    \centering
    \includegraphics[width=0.65\textwidth]{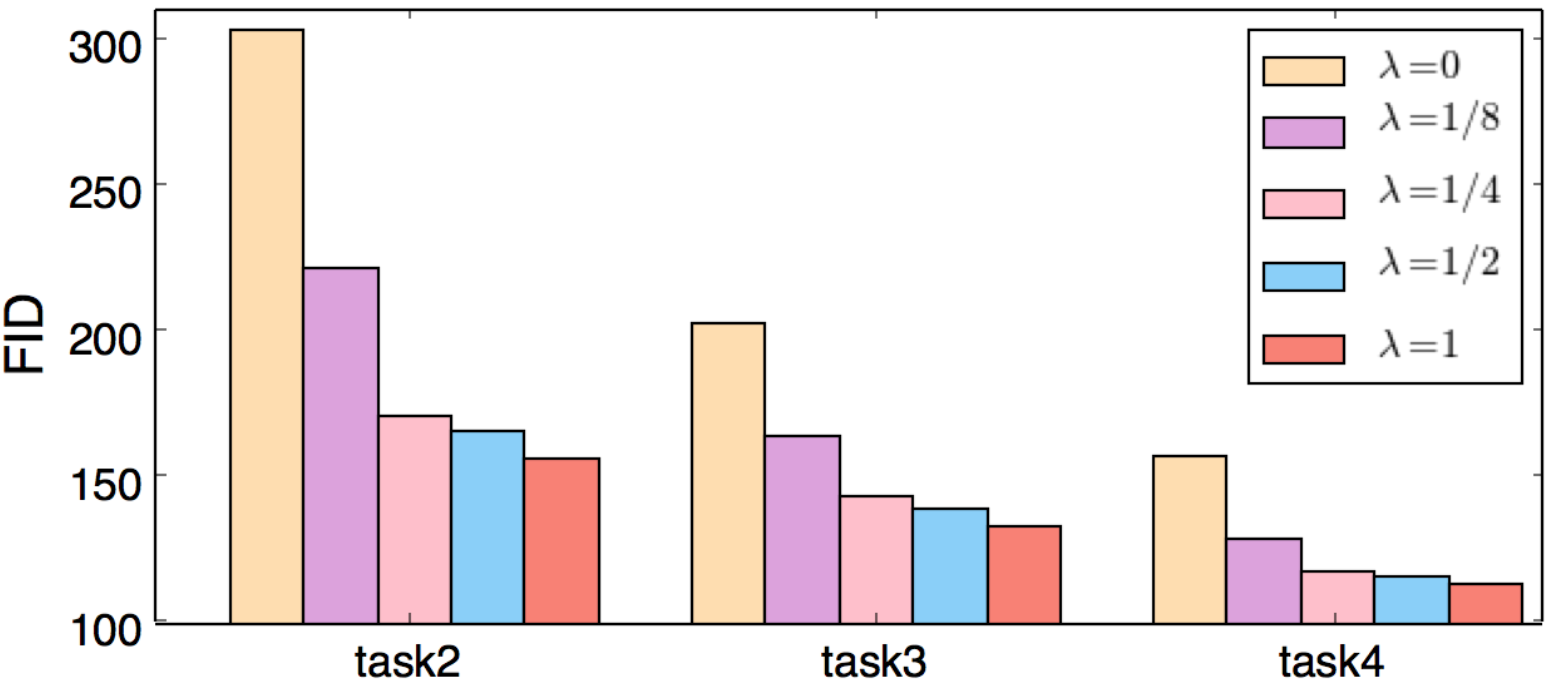}
    \caption{FID for different $\lambda$. Ablation study of choice of different values of $\lambda$. For all upcoming new tasks, the performance improves as $\lambda$ increases. However, while parameter size scales almost linearly with $\lambda$, the improvement of performance slows down gradually.}
    \label{fig:lambda}
\end{figure}

\noindent \textbf{Ablation study on choice of $\lambda$.} First we conduct an ablation study on the choice of different values of $\lambda$. For each upcoming task, we explore a set of 5 values of $\lambda$: $0$, $\frac{1}{8}$, $\frac{1}{4}$, $\frac{1}{2}$ and $1$. $\lambda=0$ corresponds to the special case where all filters are piggyback filters, and $\lambda=1$ corresponds to the special case where all filters are unconstrained filters. Figure~\ref{fig:lambda} illustrates the FID for different $\lambda$. For example, $\lambda=\frac{1}{4}$ balances well the performance and number of parameters. While parameter size scales almost linearly with $\lambda$, the improvement in performance (lower FID is better) slows down gradually.

\noindent \textbf{Ablation study on model components.} We also conduct an ablation study on the piggyback filters and unconstrained filters by comparing Piggyback GAN with the baseline models \textit{Full}, \textit{$\frac{1}{4}$Full} and \textit{Pure Factorization (PF)} on Task 2 (\textit{maps} $\rightarrow$ \textit{aerial photos}) given the same model trained on Task 1 (\textit{semantic labels} $\rightarrow$ \textit{street photos}). 

The quantitative evaluations of all approaches are summarized in Table~\ref{table:ablation-paired}. The baseline \textit{$\frac{1}{4}$Full} produces the worst classification accuracy. Since the details like building blocks are hardly visible, the classifier sometimes mistakes category \textit{facades} for category \textit{maps}.  The baseline \textit{PF} generates images that contain more details as compared to the images generated using \textit{$\frac{1}{4}$Full}. This suggests that \textit{piggyback filters} can provide valuable information of the patterns in the training data. High \textit{FID} scores for baselines \textit{$\frac{1}{4}$Full} and \textit{PF} indicate that generation qualities for both approaches are poor, and it is observed that the generated images are blurry, resulting in lots of missing details, and also contain lots of artifacts for both approaches. Our Piggyback GAN is parameter efficient and produces images having similar quality as the \textit{Full} model. 

\begin{table}[h!]
\centering
\begin{tabular}{p{1.5cm}<{\centering} p{1.5cm}<{\centering} p{1.5cm}<{\centering} p{1.5cm}<{\centering} p{1.5cm}<{\centering} p{2.5cm}<{\centering}}
\toprule
 & Full & $\frac{1}{2}$Full  & $\frac{1}{4}$Full & PF & Piggyback GAN \\
\cmidrule[0.04em](lr{.2em}){1-6} 
Acc & 97.90 & 88.10 & 84.87 & 99.82 & 97.99 \\
FID & 156.23 & 189.08 & 285.71 & 303.90 & 171.04 \\ 
\bottomrule
\end{tabular}
\caption{Model components. Ablation study of model components on paired image-conditioned generation tasks. Different models are trained and evaluated on task 2, FID and classification score on task 2 is reported.}
\label{table:ablation-paired}
\end{table}

\noindent \textbf{Generalization ability.} To demonstrate the generalization ability of our model, we learn the same $T_2$: \textit{maps} $\rightarrow$ \textit{aerial photos} with three different initial tasks $T_1$: \textit{semantic labels} $\rightarrow$ \textit{street photos}, \textit{segmentations} $\rightarrow$ \textit{facades}, and \textit{edges} $\rightarrow$ \textit{handbag photos} to evaluate whether Piggyback GAN could generalize well. The quantitative evaluation of all approaches are summarized in Table~\ref{table:generality}. The experimental results indicate that Piggyback GAN performs stable. The consistent Acc and FID scores indicate that the generated images on task $T_2$ from all three initial tasks $T_1$ have similar high image quality.

\begin{table}[h!]
\centering
\begin{tabular}{p{1.2cm}<{\centering} p{2.5cm}<{\centering} p{2.2cm}<{\centering} p{1.5cm}<{\centering}}
\toprule
$T_1$ & \makecell{\textit{semantic labels} \\ $\downarrow$ \\ \textit{street photos}} & \makecell{\textit{segmentations} \\ $\downarrow$ \\ \textit{facades}} & \makecell{\textit{edges} \\ $\downarrow$ \\ \textit{handbags}} \\ 
\cmidrule[0.04em](lr{.2em}){1-4} 
Acc & 97.99 & 97.08 & 98.09  \\ 
FID & 171.04 & 174.67 & 166.20  \\ 
\bottomrule
\end{tabular}
\caption{Generalizability. Given different models trained for different initial tasks $T_1$, same new task $T_2$ is learnt. The consistent Acc and FID scores indicate that our model generalize well and perform consistently well given different set of filter banks achieved from different tasks.}
\label{table:generality}
\end{table}

\noindent \textbf{Comparison with SOTA method and baselines.} We compare Piggyback GAN with the state-of-the-art approach~\cite{mallyaetal2018}, \textit{Sequential Fine-tuning (SFT)} and baseline \textit{Full}. Baseline \textit{Full} can be considered as the ``upper bound" approach, which provides the best performance that Piggyback GAN and \cite{mallyaetal2018} could achieve. For both Piggyback GAN and sequential fine-tuning, the model of \textit{Task2} is initialized from the same model trained on \textit{Task1}. The same model also serves as the backbone network for approach~\cite{mallyaetal2018}. 

\begin{figure*}[h!]
\begin{center}
  \includegraphics[width=\textwidth]{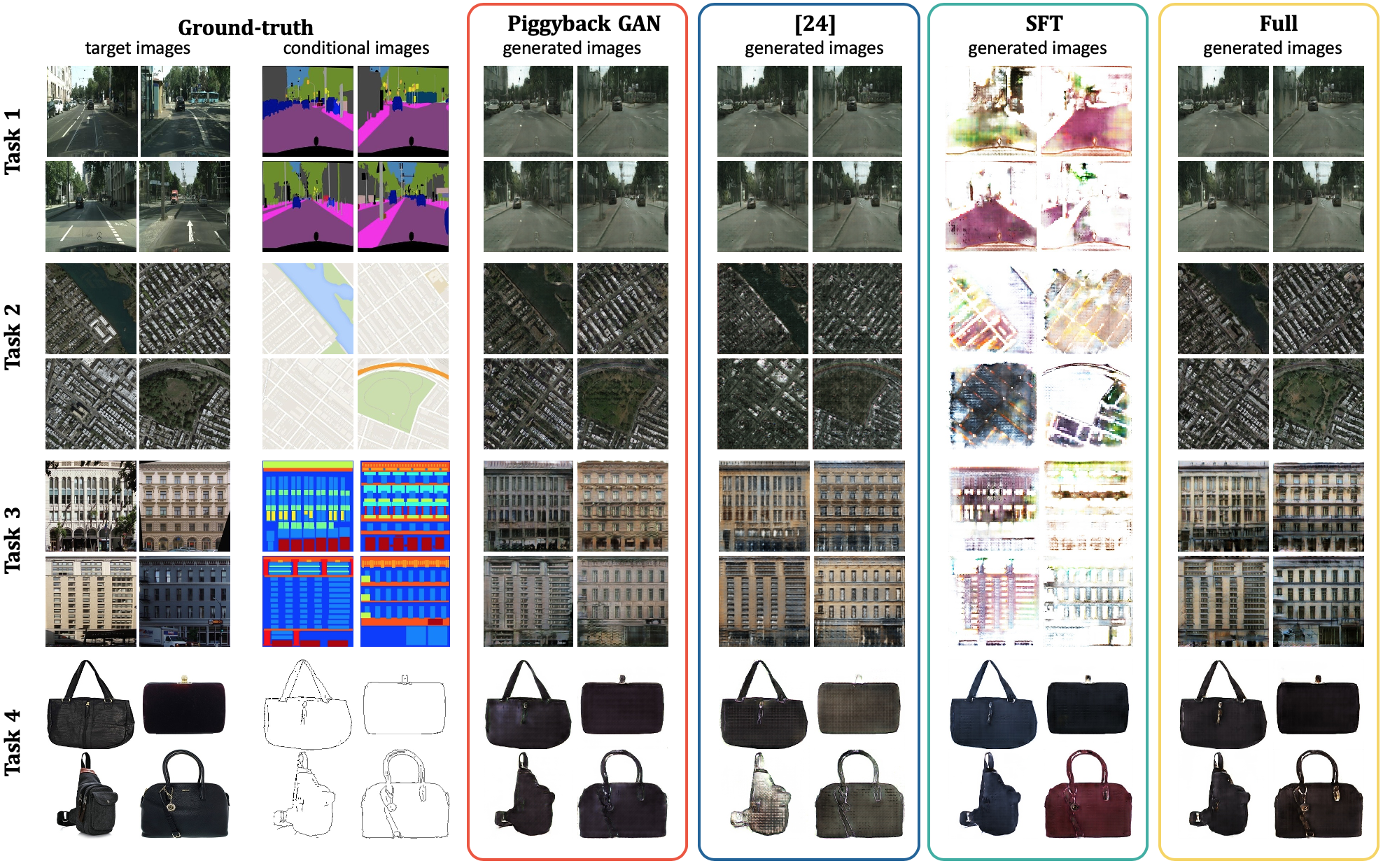}
\end{center}
\caption{Comparison among different approaches for continual learning of paired image-conditioned generation tasks. Piggyback GAN can learn the current task without forgetting the previous ones, and preserve the generation quality while shrinking the number of parameters by ``piggybacking" on an existing model.}
\label{fig:paired}
\end{figure*}

Generated images of each task for all approaches are shown in Figure~\ref{fig:paired} and the quantitative evaluations of all approaches are summarized in Table~\ref{table:comparisons-paired}. It is clear that the sequentially fine-tuned model completely forgets the previous task and can only generate incoherent edges2handbags-like (\textit{edges} $\rightarrow$ \textit{handbag photos})-like patterns. The classification accuracy suggests that state-of-the-art approach~\cite{mallyaetal2018} is able to adapt to the new task. However, it fails to capture the fine details of the new task, thus resulting in large artifacts in the generated images, especially for task 2 (\textit{maps} $\rightarrow$ \textit{aerial photos}) and task 4 (\textit{edges} $\rightarrow$ \textit{handbag photos}). In contrast, Piggyback GAN learns the current generative task while remembering the previous task and can better preserve the quality of generated images while shrinking the number of parameters by reusing the filters from previous tasks.

\begin{table}[h!]
\centering
\begin{tabular}{p{1.5cm}<{\centering} p{2.5cm}<{\centering} p{1.5cm}<{\centering} p{1.5cm}<{\centering} p{1.5cm}<{\centering}}
\toprule
 & Piggyback GAN & \cite{mallyaetal2018} & SFT & Full \\
\cmidrule[0.04em](lr{.2em}){1-5} 
Acc & 89.80 & 88.09 & 24.92 & 91.10\\
FID & 137.87 & 178.16 & 259.76 & 130.71 \\
\bottomrule
\end{tabular}
\caption{Quantitative evaluation among different approaches for continual learning of paired image-conditioned generation tasks. Different models are trained and evaluated on tasks 1-4 based on the same model trained on task 1. The average FID score over 4 tasks is reported.}
\label{table:comparisons-paired}
\end{table}

For paired image generation, we also compare our approach with \textit{Lifelong GAN}~\cite{zhai2019lifelong} and \textit{Progressive Network (PN)}~\cite{rusuetal2016}. The FID of \textit{PN} over all tasks is $141.29$. \textit{Lifelong GAN} suffers from quality degradation: the FID on cityscapes increases from 122.52 to 160.28 then to 221.07 as tasks are added, while Piggyback GAN ensures no quality degradation since the filters for previous tasks are not altered. These results show the effectiveness and the flexibility of combining the two types of filters of Piggyback GAN.

\subsection{Unpaired Image-conditioned Generation}

We also apply Piggyback GAN to another challenging scenario: unpaired image-conditioned generation $\mathrm{domain} \, A \rightarrow \mathrm{domain} \, B$. The model is trained on unpaired data, where correspondence between domain $A$ and domain $B$ does not exist, namely for each image in domain $A$ there is no corresponding ground-truth image in domain $B$. We apply our model on 2 sequences of tasks: tasks of image-to-image translation and tasks of style transfer.

\noindent \textbf{Tasks of image-to-image translation.} We convert the same 4 tasks from the paired scenario to the unpaired scenario, and compared our approach with \textit{Full}, the state-of-the-art approach~\cite{mallyaetal2018} and \textit{Sequential Fine-tuning (SFT)}. The results are shown in Table~\ref{table:comparisons-unpaired}. We observed that the Sequential Fine-tuning (SFT) cannot remember previous tasks and suffers catastrophic forgetting. Our approach produces images with high quality on par with Full model, while [24] is capable of learning each new task, however the generation quality is poor.

\noindent \textbf{Tasks of style transfer.} Two tasks in a given sequence may share the same input domain but have different output domains, e.g. $T_1$ is \textit{Photo $\rightarrow$ Monet Paintings} and $T_2$ is \textit{Photo $\rightarrow$ Ukiyo-e Paintings}. While this is not a problem when training each task separately, it does cause problems for lifelong learning. We explore a sequence of tasks of unpaired image-conditioned generation as described above. The first task is unpaired image-to-image translation of \textit{photos} $\rightarrow$ \textit{Monet paintings}, the second task is unpaired image-to-image translation of \textit{photos} $\rightarrow$ \textit{Ukiyo-e paintings}.

We compare Piggyback GAN against the state-of-the-art approach~\cite{mallyaetal2018}, \textit{Sequential Fine-tuning (SFT)} and baseline \textit{Full} which serves as the ``upper bound" approach. For both Piggyback GAN and sequential fine-tuning, the model of Task 2 is initialized from the same model trained on Task 1, which also serves as the backbone network for approach~\cite{mallyaetal2018} to allow for fair comparison. 

\begin{figure*}[h!]
\begin{center}
  \includegraphics[width=\textwidth]{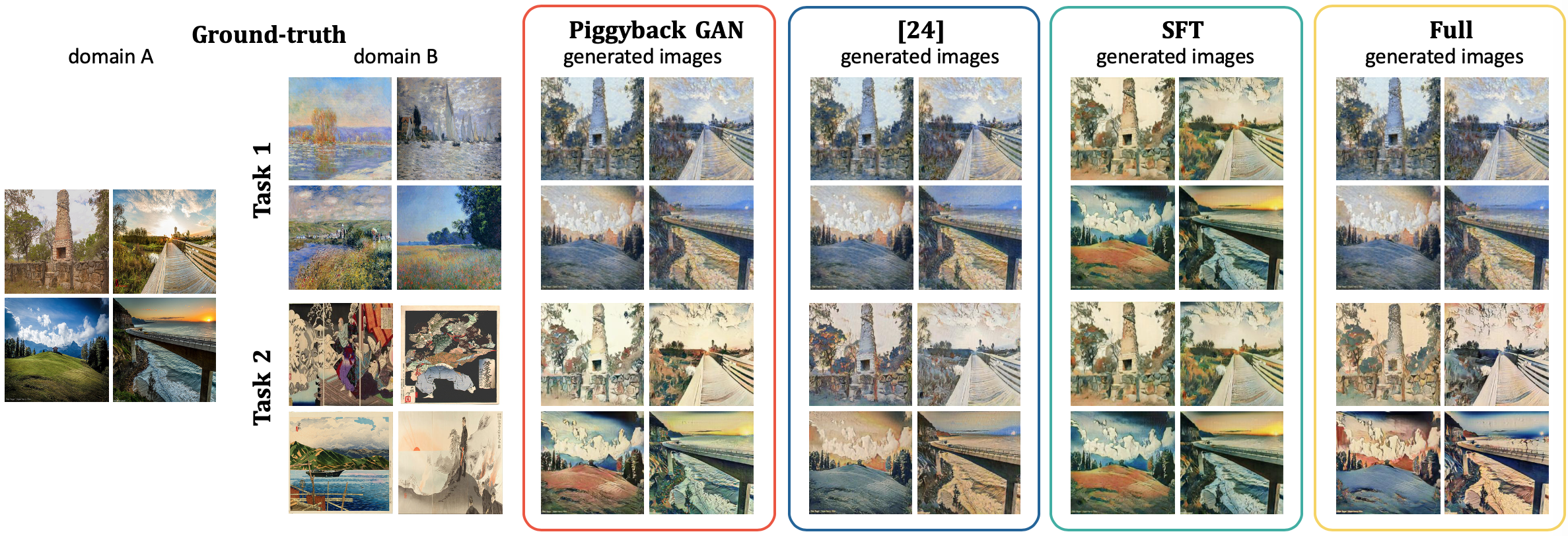}
\end{center}
\caption{Comparison among different approaches for continual learning of unpaired style transfer tasks. Piggyback GAN can preserve the generation quality the current task most without forgetting the previous ones.}
\label{fig:unpaired}
\end{figure*}

The quantitative and qualitative evaluations for comparison among different approaches are shown in Table~\ref{table:comparisons-unpaired} and Figure~\ref{fig:unpaired}, respectively. The baseline \textit{SFT} completely forgets the previous \textit{Monet} style and can only produce images in the \textit{Ukiyo-e} style. Both the state-of-the-art approach~\cite{mallyaetal2018} and Piggyback GAN are able to adapt to the new style.  The classification accuracy and FID scores indicate that Piggyback GAN can better preserve the quality of generated images, and produce images with styles most closely resembling \textit{Ukiyo-e} paintings.

\begin{table}[h!]
\centering
\begin{tabular}{p{1.5cm}<{\centering}| p{2.5cm}<{\centering} p{1.5cm}<{\centering} p{1.5cm}<{\centering} p{1.5cm}<{\centering}}
\toprule
 Tasks & \multicolumn{4}{c}{4 Image-to-image Translation Tasks}\\ 
\cmidrule[0.04em](lr{.2em}){1-5} 
 Methods & Piggyback GAN & \cite{mallyaetal2018} & SFT & Full \\
\cmidrule[0.04em](lr{.2em}){1-5} 
Acc	& 90.30	& 88.29	& 24.85	& 91.04\\
FID	& 113.89 & 149.51& 262.37 & 109.46 \\
\bottomrule
\multicolumn{5}{c}{}\\
\toprule
 Tasks & \multicolumn{4}{c}{2 Style Transfer Tasks}\\ 
\cmidrule[0.04em](lr{.2em}){1-5} 
 Methods & Piggyback GAN & \cite{mallyaetal2018} & SFT & Full \\
\cmidrule[0.04em](lr{.2em}){1-5} 
Acc & 77.97 & 70.58 & 50.00 & 78.97 \\ 
FID & 106.95 & 118.26 & 135.05 & 101.09 \\
\bottomrule
\end{tabular}
\caption{Unpaired image conditioned generation tasks. Different models are trained on all tasks based on the same model trained on task 1. The average score over all tasks is reported.}
\label{table:comparisons-unpaired}
\end{table}

% \begin{table}[h!]
% \centering
% \begin{tabular}{p{1.5cm}<{\centering} p{2.5cm}<{\centering} p{1.5cm}<{\centering} p{1.5cm}<{\centering} p{1.5cm}<{\centering}}
% \toprule
%  & Piggyback GAN & \cite{mallyaetal2018} & SFT & Full \\
% \cmidrule[0.04em](lr{.2em}){1-5} 
% Acc	& 90.30	& 88.29	& 24.85	& 91.04\\
% FID	& 113.89 & 149.51& 262.37 & 109.46 \\
% \bottomrule
% \end{tabular}
% \caption{Unpaired image-to-image translation tasks. Different models are trained and evaluated on tasks 1-4 based on the same model trained on task 1. The average FID score over 4 tasks is reported.}
% \label{table:unpaired-I2I}
% \end{table}

% \begin{table}[h!]
% \centering
% % \begin{tabular}{c c c c c }
% \begin{tabular}{p{1.5cm}<{\centering} p{2.5cm}<{\centering} p{1.5cm}<{\centering} p{1.5cm}<{\centering} p{1.5cm}<{\centering}}
% \toprule
%  & Piggyback GAN & \cite{mallyaetal2018} & SFT & Full \\
% \cmidrule[0.04em](lr{.2em}){1-5} 
% Acc & 77.97 & 70.58 & 50.00 & 78.97 \\ 
% FID & 106.95 & 118.26 & 135.05 & 101.09 \\
% \bottomrule
% \end{tabular}
% \caption{Unpaired style transfer tasks. Different models are trained and evaluated on tasks 1-2 based on the same model trained on task 1. The average FID score over 2 tasks is reported.}
% \label{table:comparisons-unpaired}
% \end{table}

\section{Conclusion}
We proposed Piggyback GAN for lifelong learning of generative networks, which can handle various conditional generation tasks across different domains. Compared to the naive approach of training a separate standalone model for each task, our approach is parameter efficient since it learns to perform new tasks by making use of the model trained on previous tasks and combining with a small portion of unconstrained filters. At the same time, our model is able to maintain image generation quality comparable to the single standalone model for each task. Since the filters learned for previous tasks are preserved, our model is capable of preserving the exact generation quality of previous tasks. We validated our approach on various image-conditioned generation tasks across different domains, and the qualitative and quantitative results show our model addresses catastrophic forgetting effectively and efficiently.

\clearpage
\bibliographystyle{splncs04}
\bibliography{egbib}
\end{document}